\documentclass[10pt,twocolumn,letterpaper]{article}

\usepackage{cvpr}
\usepackage{times}
\usepackage{epsfig}
\usepackage{graphicx}
\usepackage{amsmath}
\usepackage{amssymb}
\usepackage{subfigure}
\usepackage{algorithm,algorithmic}

\usepackage[pagebackref=true,breaklinks=true,letterpaper=true,colorlinks,bookmarks=false]{hyperref}

\cvprfinalcopy % *** Uncomment this line for the final submission

\def\cvprPaperID{5394} % *** Enter the CVPR Paper ID here

\begin{document}

%%%%%%%%% TITLE
\title{Recurrent Convolution for Compact and Cost-Adjustable Neural Networks: An Empirical Study}

\author{Zhendong Zhang \& Cheolkon Jung\\
School of Electronic Engineering, Xidian University\\
%Xi'an, 710071, China\\
{\tt\small zhd.zhang.ai@gmail.com \& zhengzk@xidian.edu.cn}
}

\maketitle
%\thispagestyle{empty}

%%%%%%%%% ABSTRACT
\begin{abstract}
   Recurrent convolution (RC) shares the same convolutional kernels and unrolls them multiple steps, which is originally proposed to model time-space signals. We argue that RC can be viewed as a model compression strategy for deep convolutional neural networks. RC reduces the redundancy across layers. However, the performance of an RC network is not satisfactory if we directly unroll the same kernels multiple steps. We propose a simple yet effective variant which improves the RC networks: the batch normalization layers of an RC module are learned independently (not shared) for different unrolling steps. Moreover, we verify that RC can perform cost-adjustable inference which is achieved by varying its unrolling steps. We learn double independent BN layers for cost-adjustable RC networks, \ie independent w.r.t both the unrolling steps of current cell and upstream cell. We provide insights on why the proposed method works successfully. Experiments on both image classification and image denoise demonstrate the effectiveness of our method.
\end{abstract}

%%%%%%%%% BODY TEXT
\section{Introduction}

Deep convolution neural networks (DCNNs) have achieved ground-breaking results on a broad range of computer vision fields, such as image classification~\cite{he2016identity}, image generation~\cite{radford2016unsupervised}, image denoise~\cite{zhang2017beyond}, object detection~\cite{ren2017faster} and so on. Despite their incredible representational power for computer vision tasks, DCNNs suffer from two problems for industrial applications:

First, DCNNs require immense computational demands and memory demands. For example, ResNet-50~\cite{he2016identity} trained on ImageNet~\cite{deng2009imagenet:} requires billions of floating-point operations (FLOPs) and hundreds megabyte memory to perform inference on a single 224 $\times$ 224 image.

Second, the inference cost and the computational graph of DCNNs are fixed for a given size of images. In many cases, cost-adjustable DCNNs are required. That is DCNNs should have the ability to adjust their inference cost w.r.t the real-time resource to achieve a better trade-off between speed and accuracy~\cite{kuen2018stochastic}. Here, we list three example cases: 1) Run a model on different devices with varied computation and memory capacities; 2) Multiple models are running simultaneously on the same device. That is, the resources assigned to a given model are varied; 3) For web-based applications, user requests have peaks and valleys.

Those two problems pose a serious challenge on deploying DCNNs for industrial applications, especially when the applications run on mobile or Internet-of-Thing (IoT) devices. To address the first problem, two lines of works have been proposed to obtain more compact DCNNs: 1) compression a pre-trained DCNN by weights pruning, quantization and sharing~\cite{han2015deep, luo2018thinet} while keeping its performance as much as possible. 2) training a compact DCNN from scratch by dynamic quantization~\cite{hubara2017quantized}, designing compact architectures~\cite{zhang2018shufflenet:} and distilling knowledge from the larger one~\cite{hinton2015distilling,romero2015fitnets:}. 

A trivial solution of the second problem is training and deploying multiple DCNNs for the same task. However, it is time consuming to train. And when deploying multiple DCNNs, the burden of devices will be increased. \cite{huang2018multi-scale,lee2015deeply-supervised,teerapittayanon2016branchynet:} train multiple classifiers which branch out of intermediate network layers. \cite{anonymous2019slimmable,kim2018doubly} integrate multiple sub-networks with different width into a single network. \cite{kuen2018stochastic} changes the spatial resolution of feature maps at varied depth.

In this paper, we argue that using recurrent convolution (RC) can obtain both compact and cost-adjustable DCNNs. RC shares the same convolutional kernels and unrolls them multiple steps, which is originally proposed to model time-space signals. Space information is captured by convolutions and time information is captured by unrolling the same convolutions. For image level tasks, there are no explicit sequential inputs. Thus it seems RC is unnecessary. However, RC has its own rule for image level tasks: suppose there is an RC network with $n$ RC layers each of which unrolls $k$ steps, then we say its depth is $nk$. If the performance of this network can match the performance of a standard DCNN with $nk$ layers (other conditions are the same), we could say the standard one is compressed by a factor $k$. In another word, the RC one is $k$ times more compact than the standard one. RC networks can be further compressed by other techniques which are complementary to RC. Moreover, if we train an RC network with different unrolling steps simultaneously, then a cost-adjustable RC network is obtained. That is, we can choose the unrolling steps of that network when performing inference.

However, the performance is not satisfactory or even worse when we directly unroll each cell of an RC network multiple steps. We believe this is caused by batch normalization~\cite{Ioffe2015Batch} (BN) layer. A BN layer captures the statistics of its input. When we share the BN layers across unrolling steps, we merge those statistics over unrolling steps into a single mean and variance vector. But there are no reasons to expect the statistics over different unrolling steps are the same. Thus, sharing the BN layers and merging the statistics would hurt the performance of an RC network. We solve this problem by a simple way: we learn independent BN layers at each unrolling step, \ie the number of BN layers is proportional to the unrolling step. Moreover, when we train a cost-adjustable RC network, we assign different groups of BN layers to a cell. Each BN group corresponds to a given step of its upstream cell. We call this double independent BN, \ie independent w.r.t both the steps of the current cell and the upstream cell. We will describe our method and the insights why it works in detail in section~\ref{sec_3} and~\ref{sec_4}. We highlight the novelties of this work as follows:
\begin{itemize}
\item Recurrent convolution has been proposed very early and many works have used it to solve computer vision tasks. However, to our knowledge, this is the first work which explicitly views RC as a network compression strategy and the first work which compares RC networks with their corresponding standard ones (with exactly the same computational graph) strictly.
\item We train an RC network with independent BN layers instead of the shared ones. This significantly improves its performance, thus indirectly plays a role in network compression. From a compression point of view, RC reduces the redundancy across layers, which is ignored by most direct compression methods.
\item We train an RC network with different unrolling steps simultaneously with double independent BN layers. This makes the RC network cost-adjustable for inference.
\end{itemize}

Experiments on both image classification (semantic level) and image denoise (pixel level) demonstrate the effectiveness of our method. The purpose of this paper is to show the potential of RC for learning compact and cost-adjustable neural networks.

\begin{figure*}[!tp]
\label{fig_method}
	\centering
 	\subfigure[Shared BN]{
 	\label{fig_BN_share}
 		\includegraphics[width=0.45\textwidth]{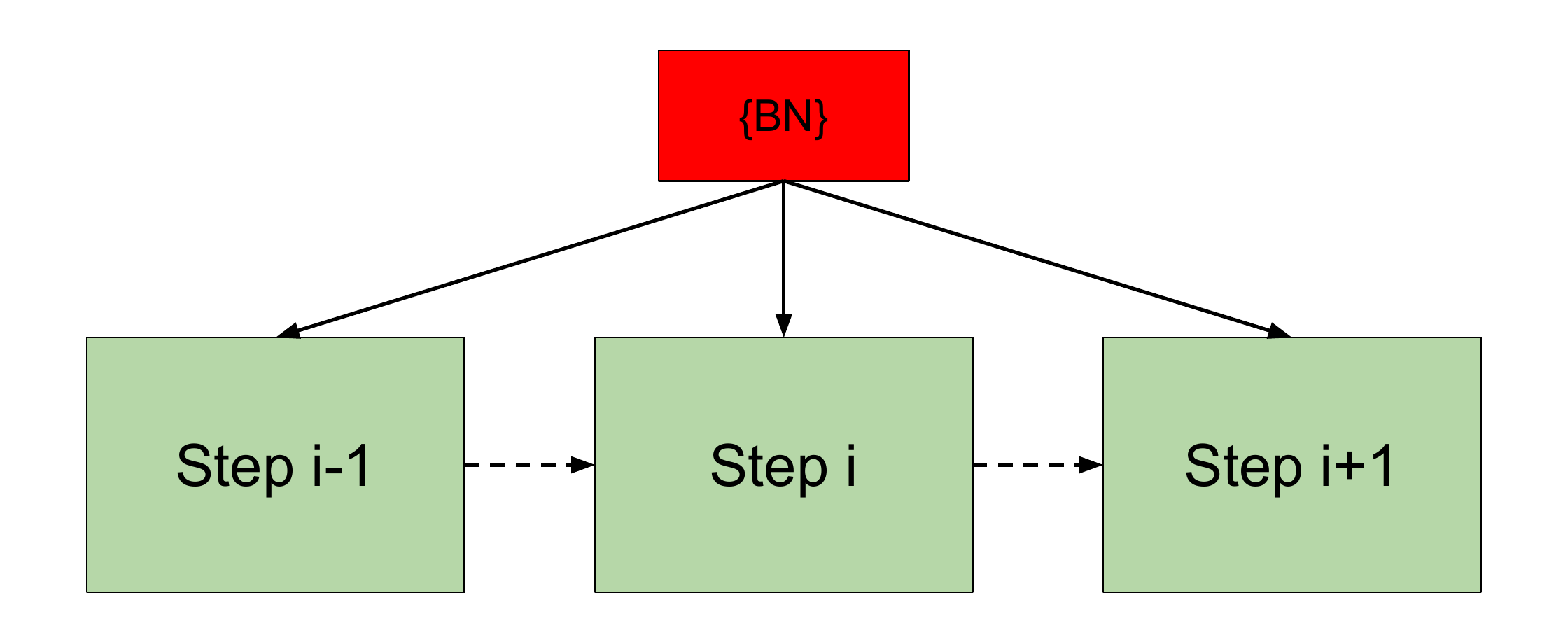}}
 	\subfigure[Independent BN]{
 	\label{fig_BN_ind}
 		 \includegraphics[width=0.45\textwidth]{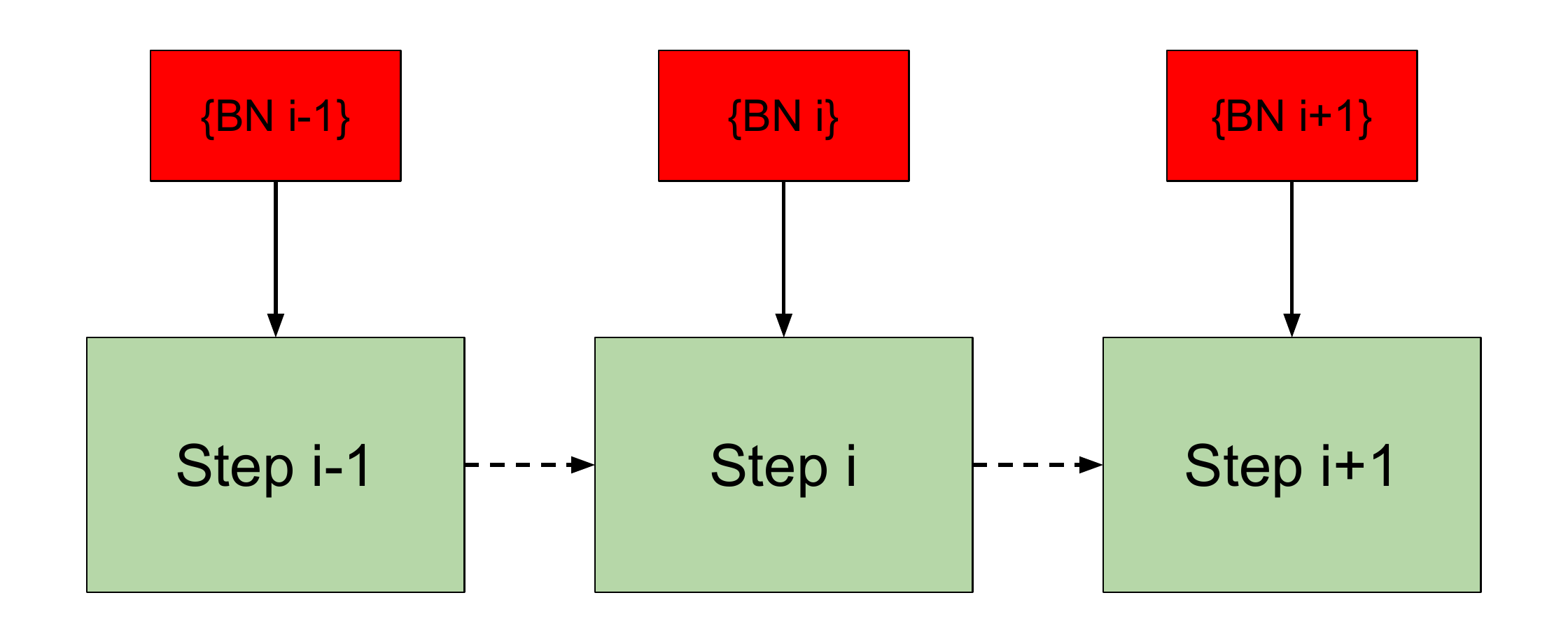}}
 	\subfigure[Double independent BN]{
 	\label{fig_BN_any}
 		  	\includegraphics[width=0.9\textwidth]{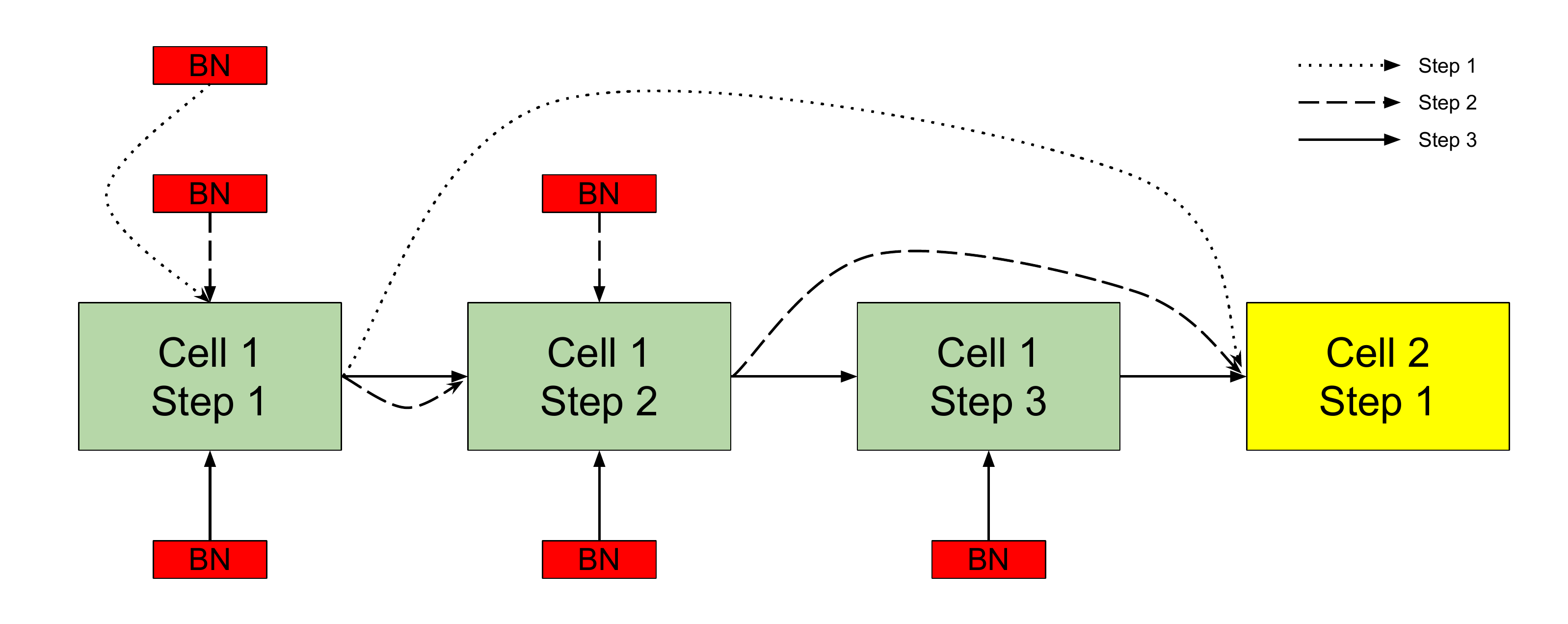}}
 \caption{Different usages of BN are compared in this figure. Shared BN is shown in (a). Independent BN is shown in (b). See main text in section~\ref{sec_31} for more information. Double independent BN for cost-adjustable RC networks is shown in (c). The unified unrolling step of each cell in (c) ranges from 1 to 3. Different types of lines show the computational path for different unrolling steps. All BN groups in (c) are independent. See main text in section~\ref{sec_4} for details.}
\end{figure*}

\section{Related Works}
\label{sec_2}
\textbf{Compact Neural Networks:} Many works have been proposed to obtain compact neural networks. Han \etal~\cite{han2015deep} compress the parameters of a network by combining weight pruning, k-means clustering, and Huffman coding. Their approach is further optimized by~\cite{choi2016towards} using Hessian-weighted k-means clustering.
Each weight is pruned independently in these two works. It is observed in~\cite{wen2016learning} that the practical acceleration is very limited due to the non-structured pruning. Luo \etal~\cite{luo2017thinet} propose ThiNet which performs filter level pruning. Thus both the number of parameters and the computational cost are reduced. However, filters are pruned at each layer greedily. The redundancy across layers is not considered. Lin \etal~\cite{NIPS2017_6813} use reinforcement learning method to train an agency which skips (equivalent to pruning) some layers. After pruning, all of those methods require fine-tuning the pruned networks. 

Another line of works trains a compact network from scratch. Zhang \etal~\cite{zhang2018shufflenet:} design ShuffleNet for mobile devices which groups filters at each layer and shuffles the order of channels of features at particular layers. Hinton \etal~\cite{hinton2015distilling} improve the performance of student (small) networks by imitating the probabilistic outputs of teacher (large) networks, which is called knowledge distilling. Further, Romero \etal~\cite{romero2015fitnets:} propose FitNets which imitates the intermediate representations learned by the teachers. The numbers of filters of students and teachers are not necessary to be equal.

For the first line of works, RC can be viewed as pruning the parameters of the whole layers. For the second line of works, RC can be viewed as a compact architecture with the same depth but fewer parameters.

\textbf{Cost-adjustable Inference:} A network that can perform inference at different computational costs depending on the user requirements, is considered to be capable of cost-adjustable inference~\cite{kuen2018stochastic}. One popular way for cost-adjustable inference is to train multiple classifiers at intermediate layers~\cite{huang2018multi-scale,lee2015deeply-supervised,teerapittayanon2016branchynet:}. One can manually set the stop point based on certain resource constraints. Or one can decide whether to stop based on the response of intermediate classifiers. The spirits of those works are similar to ours, \ie changing the computational depth of networks. However, the higher layers are not used in these works if we stop at an early layer. Our work doesn't suffer from this issue due to its recurrent nature and only a single classifier is trained. 

Instead of reducing the computational depth, Kuen  \etal~\cite{kuen2018stochastic} achieve cost-adjustable inference by dynamically changing the depth of down-sampling operators. Very recently, \cite{anonymous2019slimmable} (anonymous) integrates multiple sub-networks with varied width into a single network. And the BN layers of each sub-network are learned independently, which is similar to our approach. The difference is that BN layers are independent over width in their work while BN layers are  independent over unrolling steps  in our work.

\textbf{Recurrent Neural Networks (RNN):} RNN is powerful for modeling sequential signals~\cite{goodfellow2016deep}. The general state equation of RNN is:
\begin{equation}
\label{eq_state}
h_i = f_{\mathbf{w}} (h_{i-1}, x_i)
\end{equation}
where $f$ is a set of differentiable operators parameterized by $\mathbf{w}$. $h_i$ is the hidden state of $i$th step and $x_i$ is the sequential input of $i$th step. RC network is a particular kind of RNN whose sequential inputs are processed by convolutions. For image level tasks, there are no explicit sequential inputs. Thus the state equation degrades into
\begin{equation}
\label{eq_dstate}
h_i = f_{\mathbf{w}}(h_{i-1})
\end{equation}
We can simply understand $h_i$ as the feature maps of a convolutional layer.

BN is first introduced into RNN by Laurent \etal~\cite{laurent2016batch}. BN is only applied to the sequential inputs, \ie $x_i$ in their work. Then Cooijmans \etal~\cite{cooijmans2017recurrent} show it is also helpful to apply BN to the hidden states. BN layers are shared in both of these two works. While in our work, we learn independent BN layers over unrolling steps.

\section{RC for Compact Neural Networks}
\label{sec_3}
\subsection{Independent BN}
\label{sec_31}
The logic behind RC for compact neural networks is easy to understand. Suppose there is an RC network with $n$ RC layers each of which unrolls $k$ steps, then we say its depth is $nk$. If the performance of this network can match the performance of its corresponding standard one with $nk$ layers, we could say the standard one is compressed by a factor $k$. The word "corresponding" means the standard network has exactly the same computational graph compared with the RC one which is fully unrolled. RC network shares parameters of the whole layers and thus reduces the redundancy across layers. 

However, when we directly unroll each cell of an RC network multiple steps, the performance is not satisfactory and even worse than the single step network's. We believe this is caused by the shared BN layers. BN~\cite{Ioffe2015Batch} significantly improves the training speed and generalization abilities of DCNNs which is the default choice of most modern network architectures. In this work, we learn independent BN layers at each unrolling step. The number of BN layers is  proportional to the unrolling steps. See Fig.~\ref{fig_BN_share} and Fig.~\ref{fig_BN_ind} for comparing those two strategies. 

Of course, learning independent BN layers will introduce more BN layers compared with sharing them. However, both memory costs and computation costs of BN layers are only a very small proportion of the whole network's. For ResNet-50, the percentage of parameters in BN layers is only $0.21 \%$. Thus using independent BN layers almost doesn't increase the device burden.

 \begin{figure*}
  	\centering
  	\subfigure[$\mu$]{
  		\includegraphics[width=0.23\textwidth]{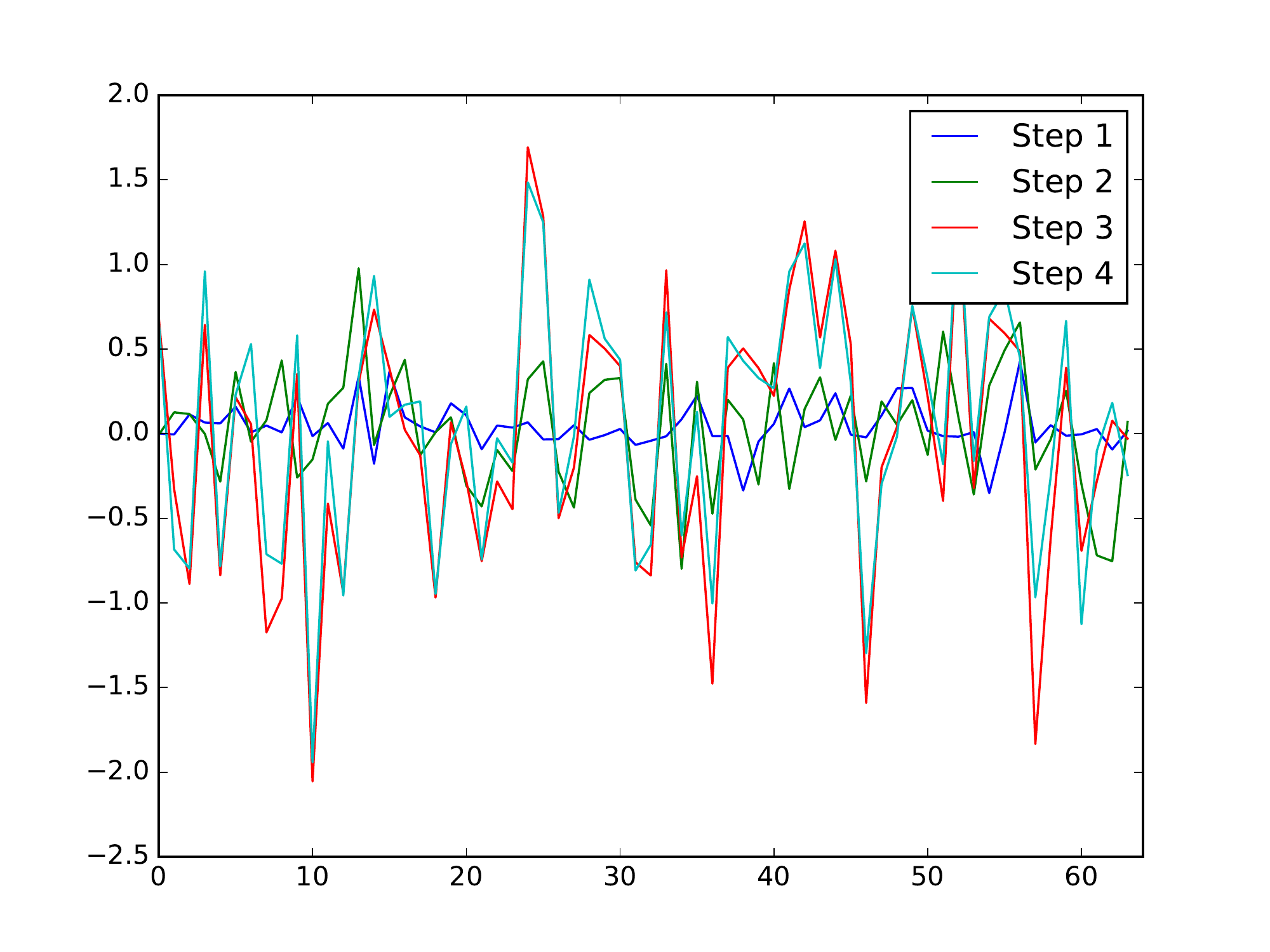}}
  	\subfigure[$\sigma^2$]{
  		\includegraphics[width=0.23\textwidth]{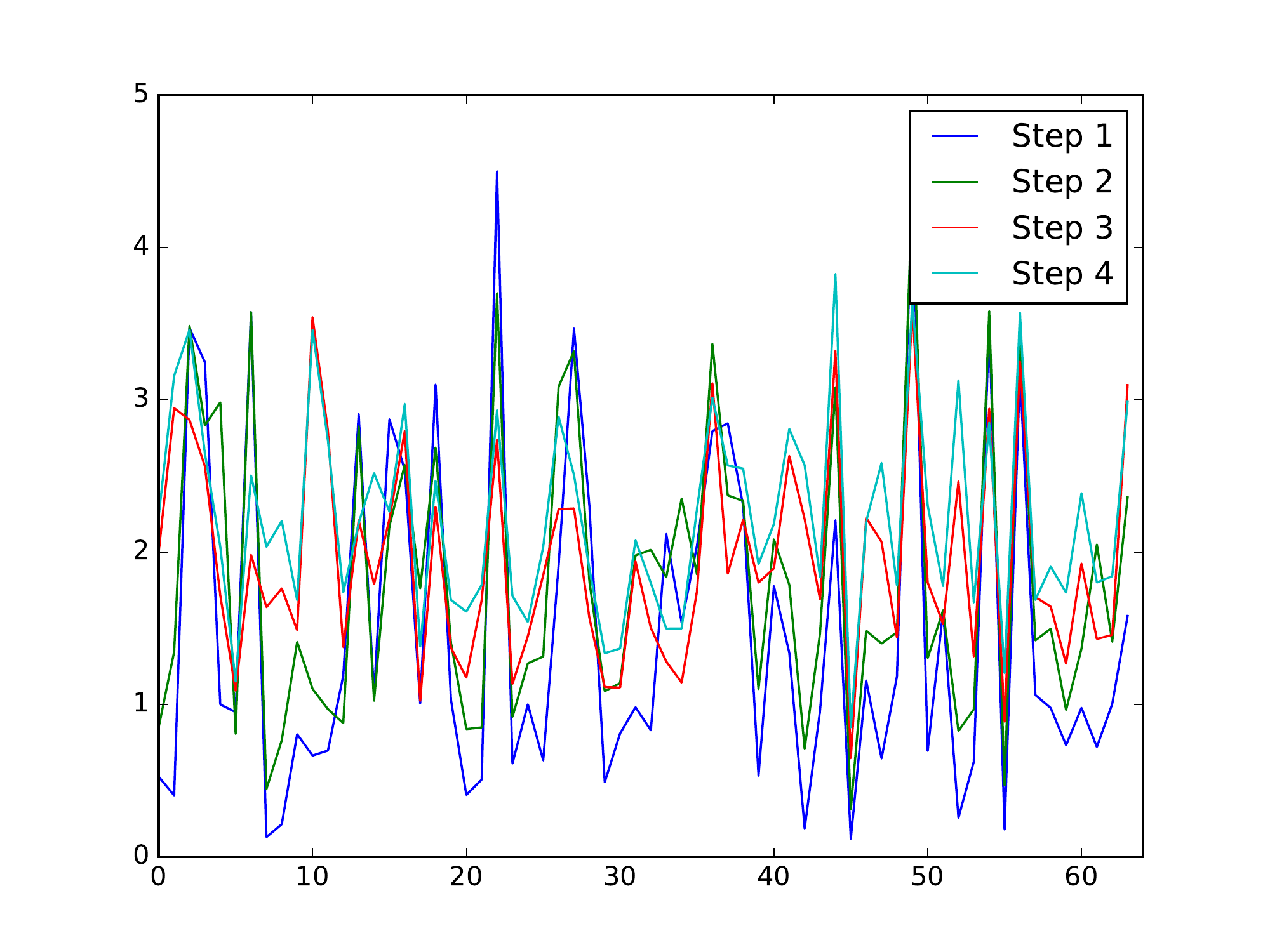}}
  	\subfigure[$\gamma$]{
  	  		\includegraphics[width=0.23\textwidth]{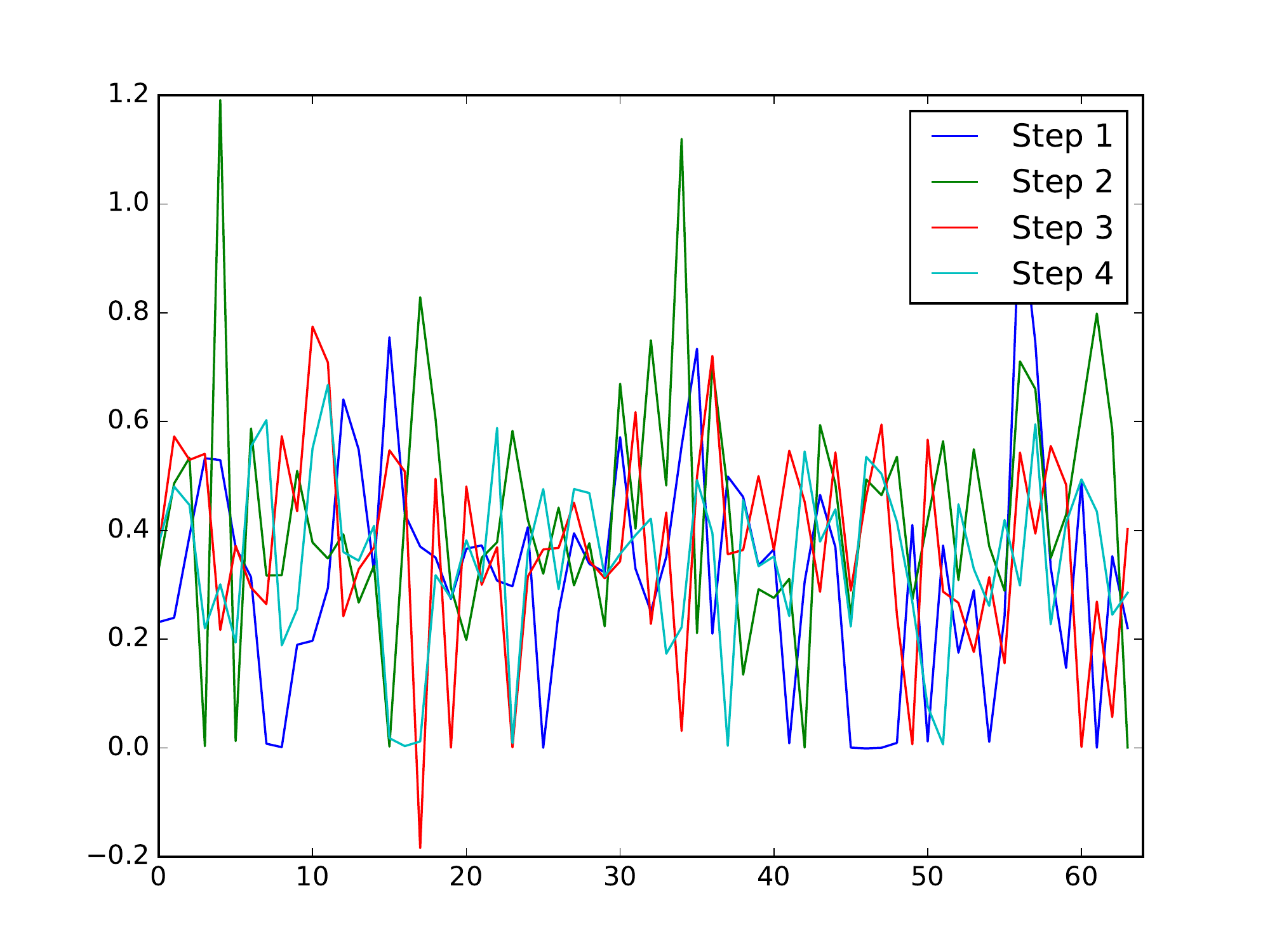}}
  	  	\subfigure[$\beta$]{
  	  		\includegraphics[width=0.23\textwidth]{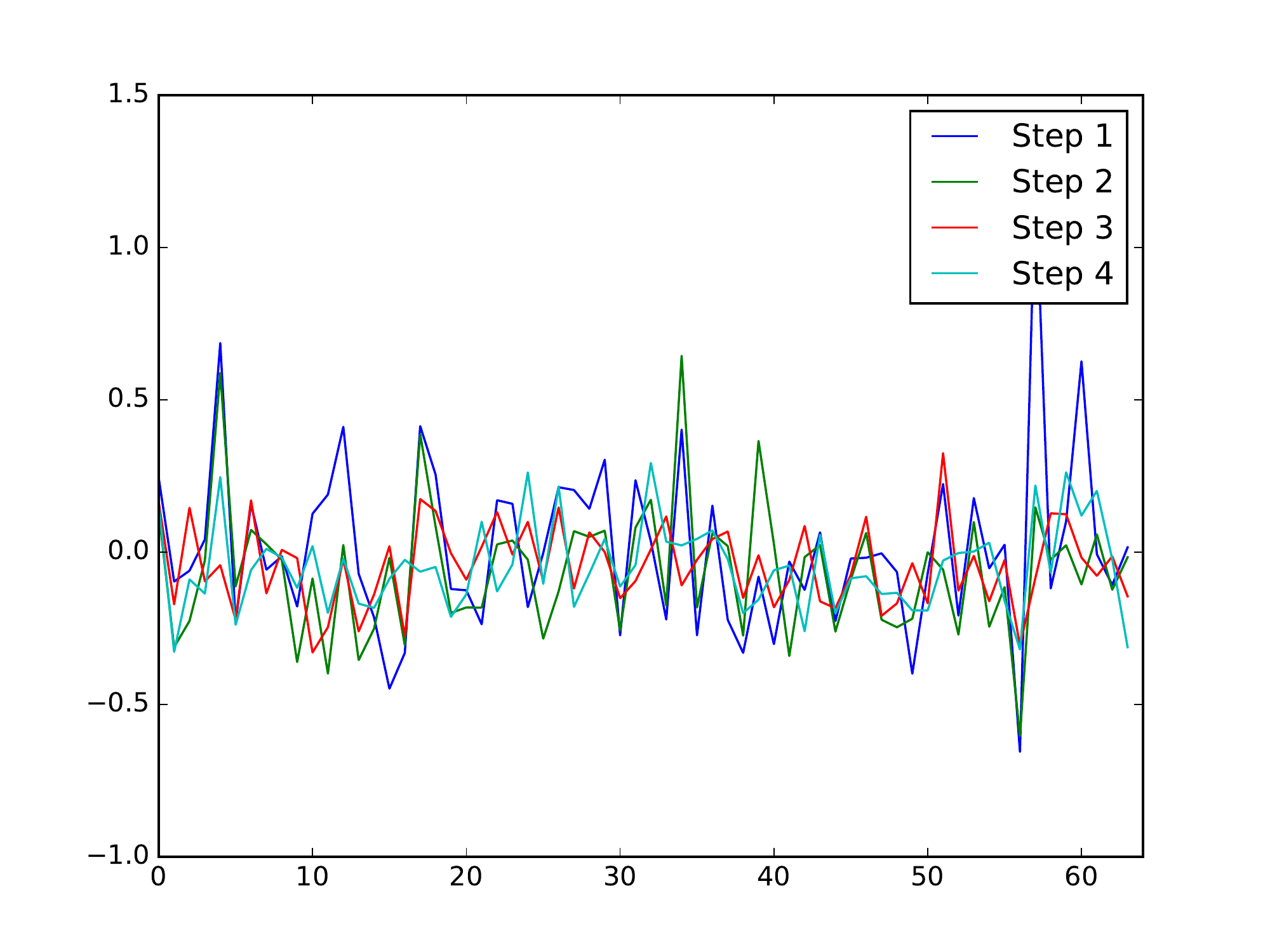}}
  	\caption{Variables of the first independent BN layer over steps trained on CIFAR-10.}
  	\label{fig_BN_0}
   \end{figure*}

 \begin{figure*}
  	\centering
  	\subfigure[$\mu$]{
  		\includegraphics[width=0.23\textwidth]{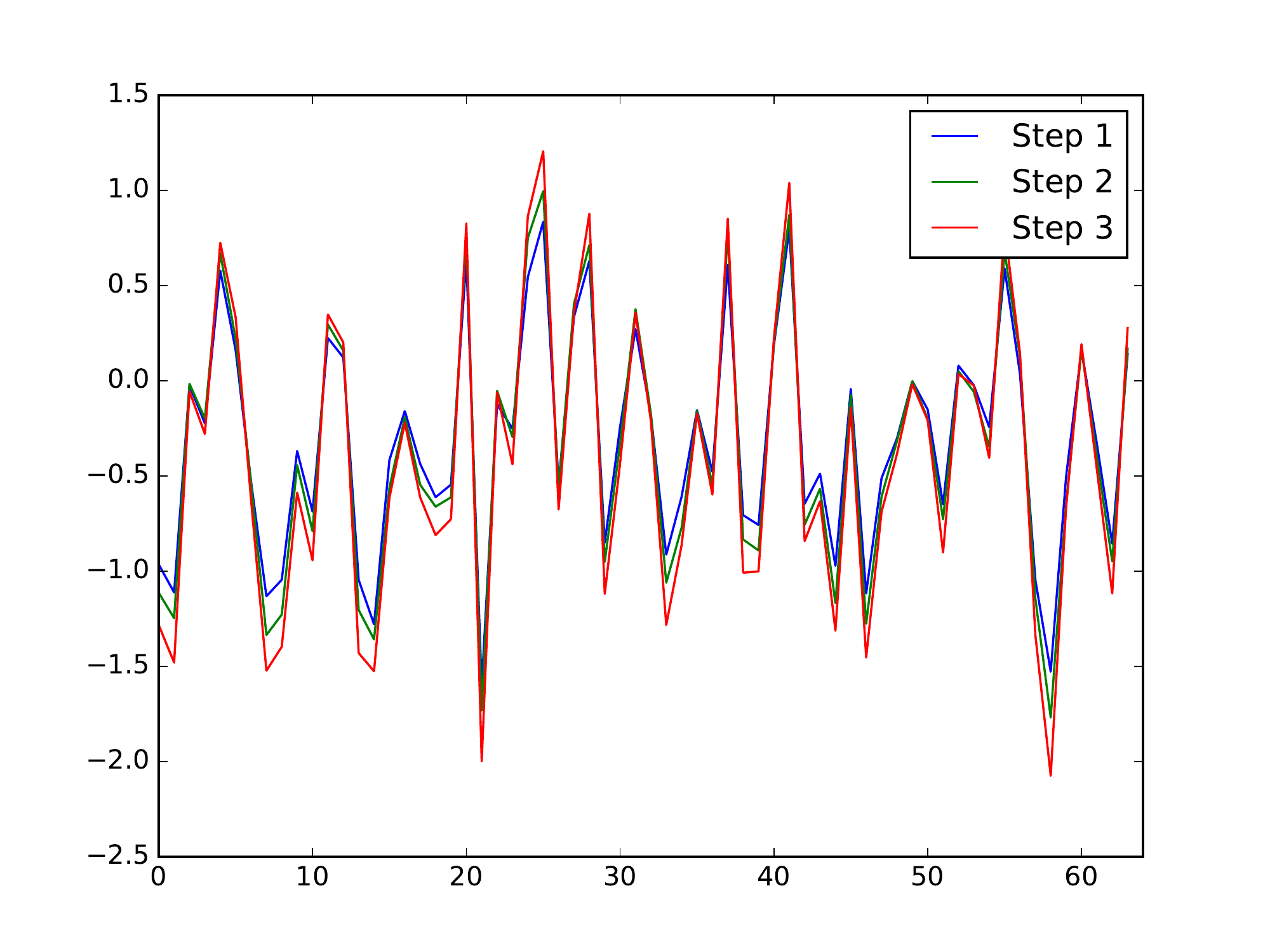}}
  	\subfigure[$\sigma^2$]{
  		\includegraphics[width=0.23\textwidth]{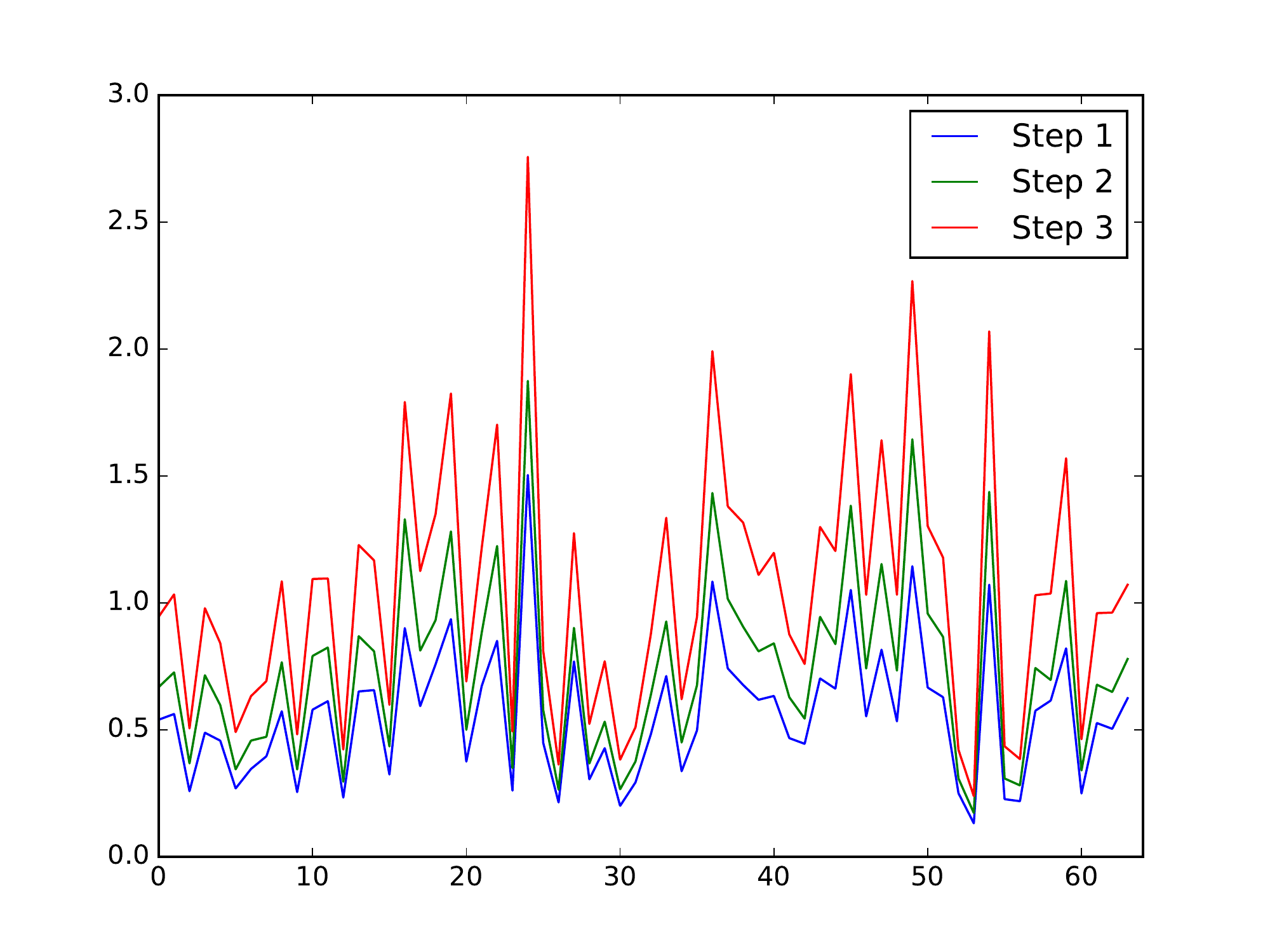}}
  	\subfigure[$\gamma$]{
  	  		\includegraphics[width=0.23\textwidth]{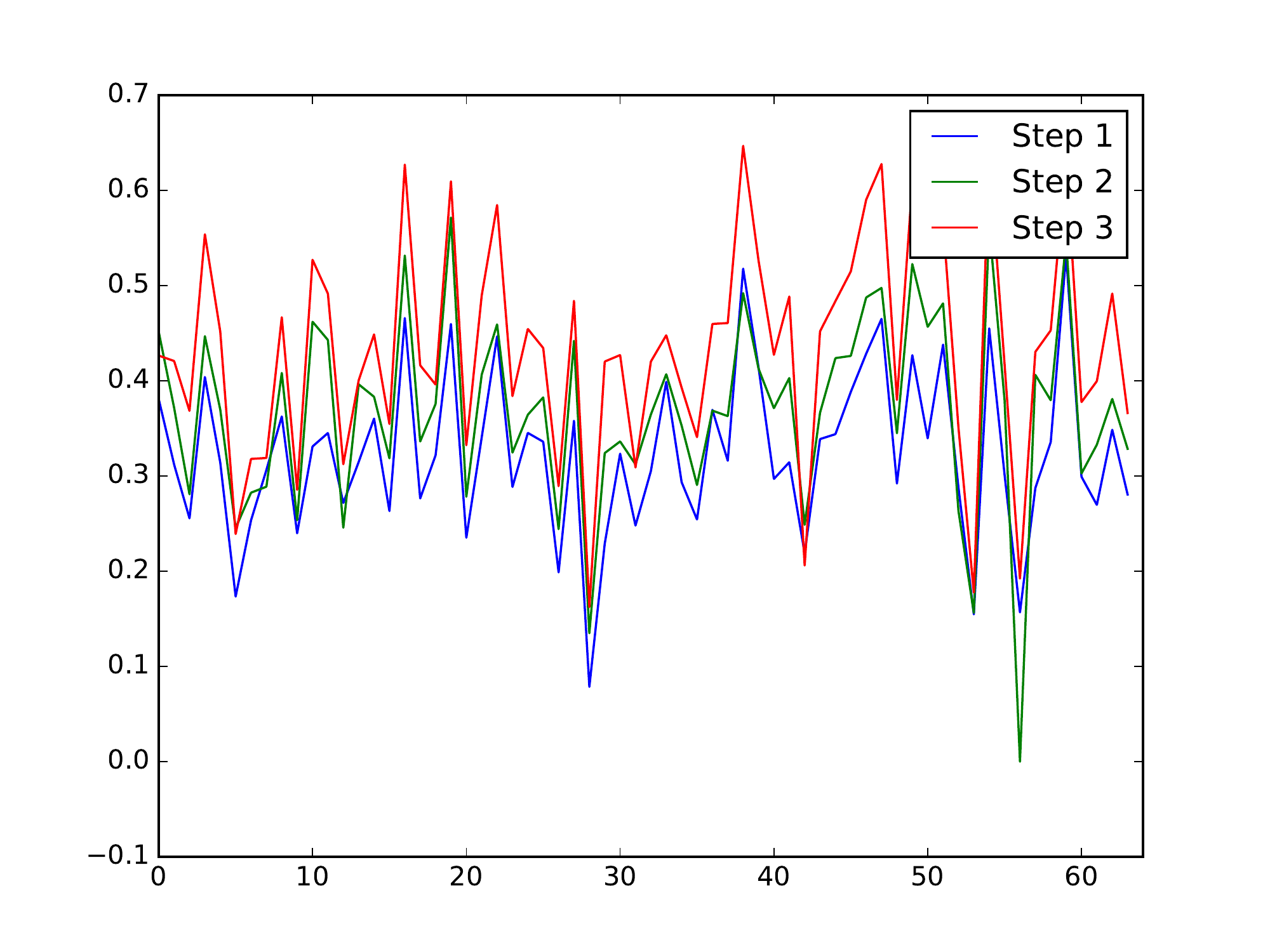}}
  	  	\subfigure[$\beta$]{
  	  		\includegraphics[width=0.23\textwidth]{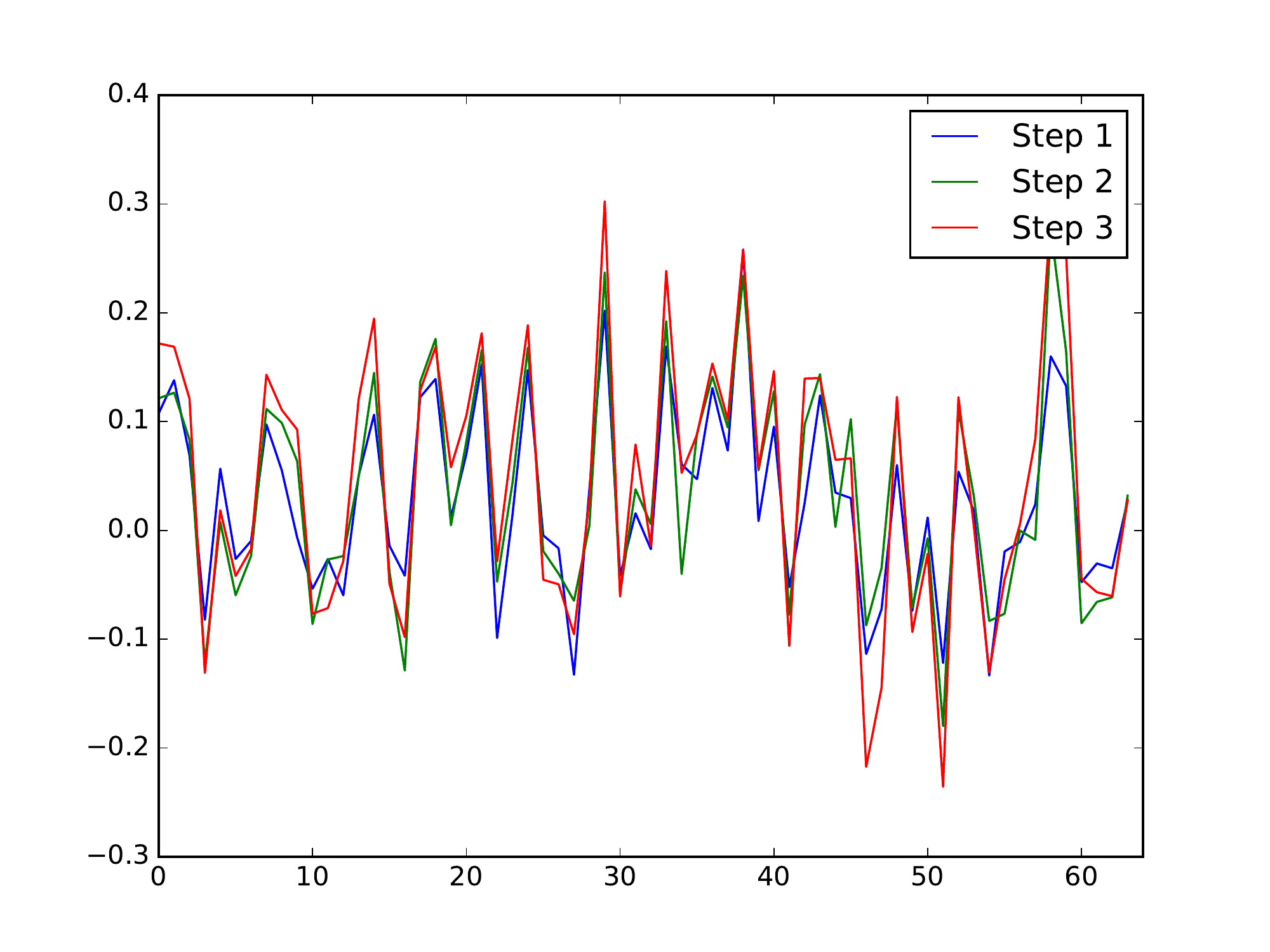}}
  	\caption{We show the learned parameters of the first BN layer of $T_2$ when $C_1$ unrolls different steps trained on CIFAR-100. See main text in section~\ref{sec_53} for details.}
  	\label{fig_BN_1}
   \end{figure*}
   
\subsection{Why Independent BN Works}
A BN layer captures the first and second order statistics of its input, then scales and shifts the normalized input by its learned parameters.
\begin{eqnarray}
\gamma \frac{x-\mu}{\sqrt{\sigma^2+\varepsilon}} + \beta
 \end{eqnarray}
where $\mu$ and $\sigma^2$ are the mean and variance of $x$ respectively. $\gamma$ and $\beta$ are the learned scale and shift parameters by stochastic gradient descent (SGD). If we share BN layers, the statistics of inputs over different unrolling steps are summarized as a single mean vector and a single variance vector. That is the information over steps are lost. Learning independent BN layers avoids lose of the information over steps. In fact, there are no reasons to expect the statistics over different unrolling steps are the same. If this is true, it is helpless to unroll a cell several steps.

Another reason is learning independent BN layers improves the representation power of an RC network.  As mentioned earlier, for image level tasks, there are no explicit sequential inputs. Thus the state equation of an RC cell is degraded into Eq.~\ref{eq_dstate}, \ie $h_i = f_{\mathbf{w}}(h_{i-1})$. The mapping function $f_{\mathbf{w}}$ is exactly the same w.r.t $h$ at each unrolling step due to the lack of sequential inputs. This limits the representation power of RC networks. If we use independent BN layers at each step, we can re-formalize its state equation as follows:
\begin{equation}
\label{eq_bnstate}
h_{i+1} = f_{\mathbf{w}_c}(h_i, \mathbf{b}_{i+1})
\end{equation}
where $\mathbf{w}_c$ denotes the shared convolutional filters and $\mathbf{b_i}$ denotes the parameters of BN layers at $i$th step. This state equation is the same as the general state equation of RNN in Eq.~\ref{eq_state}. The parameters of BN layers at each step act as the sequential inputs. Now, the mapping function w.r.t $h$ is varied at each step. In another word, the mapping function at $i$th step is conditioned by $\mathbf{b_{i+1}}$. In this way, the representation power of an RC network is improved.

\subsection{Practical Considerations}
\textbf{Choice of RC Cell:} We can choose a single layer or a group of adjacent layers as an RC cell.
Due to its recurrent nature, the numbers of input channels and output channels of an RC cell should be the same. In principle, we can change the spatial resolution of feature maps inside an RC cell. However, as we need to unroll each cell varied steps (See section~\ref{sec_4}), it is hard to control the spatial resolution if we change it inside each cell. Thus in this paper, we simply keep the size of the inputs and outputs of an RC cell fixed. 

\textbf{Training Policy:}
Denote $\mathbf{w}$ as the shared convolutional filters, $\mathcal{L}$ as the final loss and $n$ as the maximum unrolling step. Then the gradient of $\mathcal{L}$ w.r.t $\mathbf{w}$ is
\begin{equation}
\frac{\partial \mathcal{L}}{\partial \mathbf{w}} = \sum_{i=0}^{n} \frac{\partial \mathcal{L}}{\partial h_n} \frac{\partial h_n}{\partial h_i} \frac{\partial h_i}{\partial \mathbf{w}}
\end{equation}
Multiple terms of gradients are added up as the gradient of $\mathbf{w}$. Those gradient terms should have belonged to independent parameters. That is the shared parameters are updated more frequently than unshared ones. We empirically find that setting a smaller learning rate to the shared parameters achieves better results when training RC networks. Specifically, the learning rate of shared parameters is set to $50\%$ of unshared ones in this paper. Gradient clip is also helpful to stable the training process, especially when $n$ is large.

\section{RC for Cost-adjustable Inference}
\label{sec_4}
\subsection{Double Independent BN}
\label{sec_41}
A network is cost-adjustable if it can dynamically change its computational graph and cost during inference. In section~\ref{sec_3}, we train RC networks whose cells are unrolled with a fixed step during training and inference.
It seems straightforward to make an RC network cost-adjustable: we can unroll each cell with a random step at each iteration during training. The number of steps is sampled from a pre-defined discrete distribution at each iteration. Then, one can perform inference by unrolling each cell arbitrary step whose probability is not zero in the pre-defined distribution, based on users requirements.

We must consider how to use BN in such a case. Because now the unrolling step is varied but the total number of BN layers are fixed. A native way to handle this issue is that we initialize $n$ groups of BN layers for an RC cell, where $n$ is the maximum unrolling step. $i$th group corresponds to $i$th step. When the sampled step is equal to $t$ at current iteration, then only the first $t$ groups are used.

However, the performance is not satisfactory for above straightforward way. We believe this is also caused by BN. Suppose we have two RC cells, $C_{i}$ and $C_{i+1}$. The output of $C_i$ is the input of $C_{i+1}$. Because $C_i$ is unrolled with a varied step, the statistics of its output, \ie the statistics of $C_{i+1}$'s input are also varied. See Fig.~\ref{fig_BN_any} for an example. The first BN layer of $C_{i+1}$ can just match the output of $C_{i}$ with a fixed step. When we merge the statistics of $C_{i}$'s output into a single mean vector and a single variance vector, we lose its information over steps. Now, $C_{i+1}$ can't realize which step its input is generated by.  

As in Eq.~\ref{eq_bnstate}, the mapping function of each cell is conditioned on its own unrolling step if we use independent BN layers. Now each cell is also required to be conditioned on the unrolling step of its upstream cell. Then it can realize which step its input is generated by. This can be achieved by introducing more groups of BN layers for each cell.  Suppose the maximum step of $C_{i}$ is $m$ and the maximum step of $C_{i+1}$ is $n$. Then we need $mn$ groups of BN layers for $C_{i+1}$, instead of $n$ groups. Denote $\mathbf{B}$ as a matrix with size $m\times n$ whose element is a group of BN layers for $C_{i+1}$. If the step of $C_{i}$ is $t_{i}$ and the step of $C_{i+1}$ is $t_{i+1}$ at current iteration, then only the first $t_{i+1}$ elements of $t_i$th row are used. 
\begin{gather}
\begin{array}{cccc}
\label{eq_matrix}
b_{1,1} & b_{1,2} & b_{1,3} & b_{1,4} \\
\mathbf{b_{2,1}} & \mathbf{b_{2,2}} & \mathbf{b_{2,3}} & b_{2,4} \\
b_{3,1} & b_{3,2} & b_{3,3} & b_{3,4} \\
b_{4,1} & b_{4,2} & b_{4,3} & b_{4,4}
\end{array}
\end{gather}
This is an example when $ t_i=2, t_{i+1}=3$. Boldface means the element which is used. We call such strategy \emph{double independent BN}, \ie independent to both $t_i$ and $t_{i+1}$.

\begin{algorithm}
\caption{Inference}
\label{alg:A}
\begin{algorithmic}
\STATE \textbf{Require:} network architecture $N$ and its parameters,  input $x$, unrolling step $s$ and BN layers $\mathbf{B}$ for step $s$
\FOR{$module$ in $N$}
\IF{$module$ is recurrent} 
\FOR{$t=1, 2, ..., s$}
\STATE $x \leftarrow module(x, b_{t})$ 
\ENDFOR
\ELSE
\STATE $x \leftarrow module(x, b_{1})$
\ENDIF
\ENDFOR
\STATE where $b$ is the BN layers of $module$
\STATE \textbf{Return:} $x$ 
\end{algorithmic}
\end{algorithm}

\begin{algorithm}
\caption{Training}
\label{alg:B}
\begin{algorithmic}
\STATE \textbf{Require:} network architecture $N$, distribution of unrolling steps $\mathcal{S}$, BN groups $\mathbf{B}$ and loss function $\mathcal{L}$
\FOR{$i=1, 2, 3, ...$}
\STATE sample $(x, y)$ from dataset.
\STATE sample $s$ from $\mathcal{S}$
\STATE $\hat{y} \leftarrow Inference(x, s, \mathbf{B}_s)$
\STATE $loss \leftarrow \mathcal{L}(\hat{y}, y)$
\STATE back-propagate w.r.t $loss$ and update parameters
\ENDFOR
\end{algorithmic}
\end{algorithm}

\subsection{Practical Considerations}
\textbf{How to sample steps:} Suppose there are 4 RC cells each of which has a step ranging from 1 to 3. Then the computational graph of the overall network has $3^4$ possible combinations. That is the complexity of the overall network grows fast if the step of each cell is sampled independently. We empirically find it is hard to train RC networks due to so many possible combinations. 

In this paper, we set a unified unrolling step to all RC cells. Specifically, we first draw a single number $t$ for the pre-defined distribution. Then each cell unrolls $t$ steps at current iteration. Now the computational graph has at most $n$ possible combinations where $n$ is the pre-defined maximum unrolling step. We empirically find this simple variation makes RC networks much easier to train. Another advantage is that only lower triangular elements of $\mathbf{B}$ in Eq.~\ref{eq_matrix} are required. A compromise maybe also work. We find it is better to assign higher probabilities to larger unroll steps. Our intuition is that larger unrolling steps correspond to deeper networks, thus require more iterations to train.

\textbf{How to train:} The training algorithm and the inference algorithm are shown in Alg.~\ref{alg:B} and Alg.~\ref{alg:A} respectively. During training, we only sample a single step from $\mathcal{S}$. Then we perform the forward-backward loop for that step. An alternative way is that we perform inference and calculate the loss for all possible steps at each iteration. We sum over all of those loss terms weighted by the probabilities of their corresponding steps. And we update the parameters w.r.t the aggregated loss at once. By such a training strategy, the performance is improved only marginally but the training speed is decreased several times. We believe the training algorithm in Alg.~\ref{alg:B} achieves a better trade-off and we use it in all experiments.

\begin{figure*}[t]
\centering
 \includegraphics[width=0.9\textwidth]{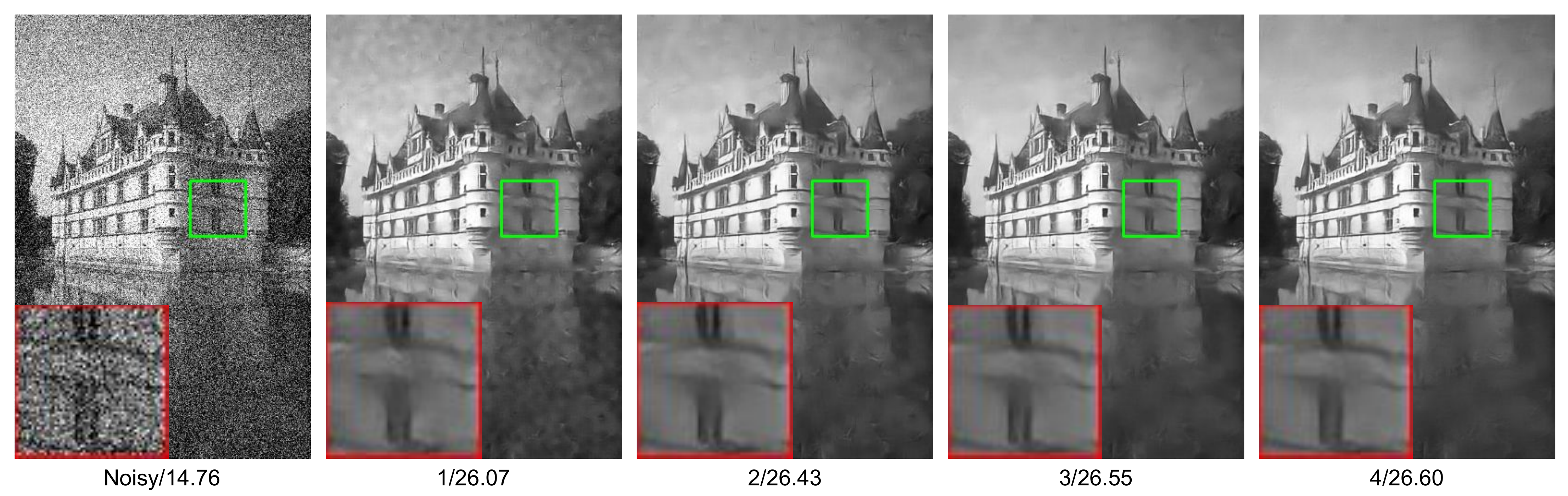}
 \caption{Denoise examples of $R_3$ with varied unrolling steps.}
 \label{fig_imgs}
\end{figure*}
\section{Experiments}
\label{sec_5}
We evaluate our method on both image classification which is a fundamental semantic level task and image denoise which is a fundamental pixel level task. For image classification, all of networks are trained with cross-entropy loss and evaluated with their error rates on validation set. CIFAR-10 and CIFAR-100~\cite{krizhevsky2009learning} are used as the dataset. Both CIFAR-10 and CIFAR-100 have 50K training samples and 10K test samples, each of which is a $32\times 32$ color image. The former has 10 classes of images while the latter has 100 classes of images. For image denoise, all networks are trained with $L_2$ loss and evaluated with their PSNR on validation set. We train denoise networks with noise level $15$, $25$ and $50$ respectively on Berkeley segmentation dataset (BSD)~\cite{roth2009fields}. This dataset has 232 training images and 68 test images. We follow experimental settings of~\cite{zhang2017beyond}. Residual learning for denoise~\cite{zhang2017beyond} is also used.
\begin{equation}
\arg \min_{\mathbf{w}} \lVert f_{\mathbf{w} }(x)+x-y \rVert_2^2
\end{equation}
where $x$ is a noisy image and $y$ is its corresponding clean image. $f_{\mathbf{w}}(x)$ is the output of the network. All networks are implemented in PyTorch~\cite{paszke2017automatic}.

We first check the effectiveness of our method by training relatively small networks with strictly controlled conditions. For image classification, we use the following architecture:
\begin{equation}
\label{eq_arch_res18}
C_1^n(64) \rightarrow InvPool \rightarrow C_2^n(256)
\end{equation}
where $C_1^n(64)$ means the first RC cell with 64 channels and is unrolled $n$ steps. $C_2^n(256)$ means the second RC cell with 256 channels and is unrolled $n$ steps. $InvPool$ is the invertible downsampling operation described in~\cite{dinh2016density}, which consists in reorganizing the initial spatial channels into the 4 spatially decimated copies obtainable by $2 \times 2$ spatial sub-sampling. 
%For example, if the size of feature maps is $20\times20\times 16$. After applying $InvPool$, it will become $10\times 10\times 64$. 
We use $InvPool$ because it is parameter free which eliminates the distraction of non-recurrent parameters. The first and last layers are omitted. We apply average pooling after $\lceil n/2\rceil$th step for each cell. Each cell is a preact-resblock~\cite{he2016identity} which contains two convolutional layers and two BN layers. For image denoise,  we use the following architecture:
\begin{equation}
\label{eq_arch_denoise}
C_1^n(64) \rightarrow C_2^n(64) \rightarrow C_3^n(64)
\end{equation}
Each cell is a composite of a convolutional layer, a BN layer, and ReLu activation.

Since the unrolling step of each cell is unitized, we denote $R_2^n$ as an RC network with 2 cells each of which unrolls $n$ steps. And we denote $S_2^n$ as a standard network whose computational graph is \emph{exactly the same} as the unrolled $R_2^n$. $n$ ranges from 1 to 4 for both Eq.~\ref{eq_arch_res18} and Eq.~\ref{eq_arch_denoise}.

 \begin{figure*}[t]
  	\centering
  	\subfigure{
  		\includegraphics[width=0.3\textwidth]{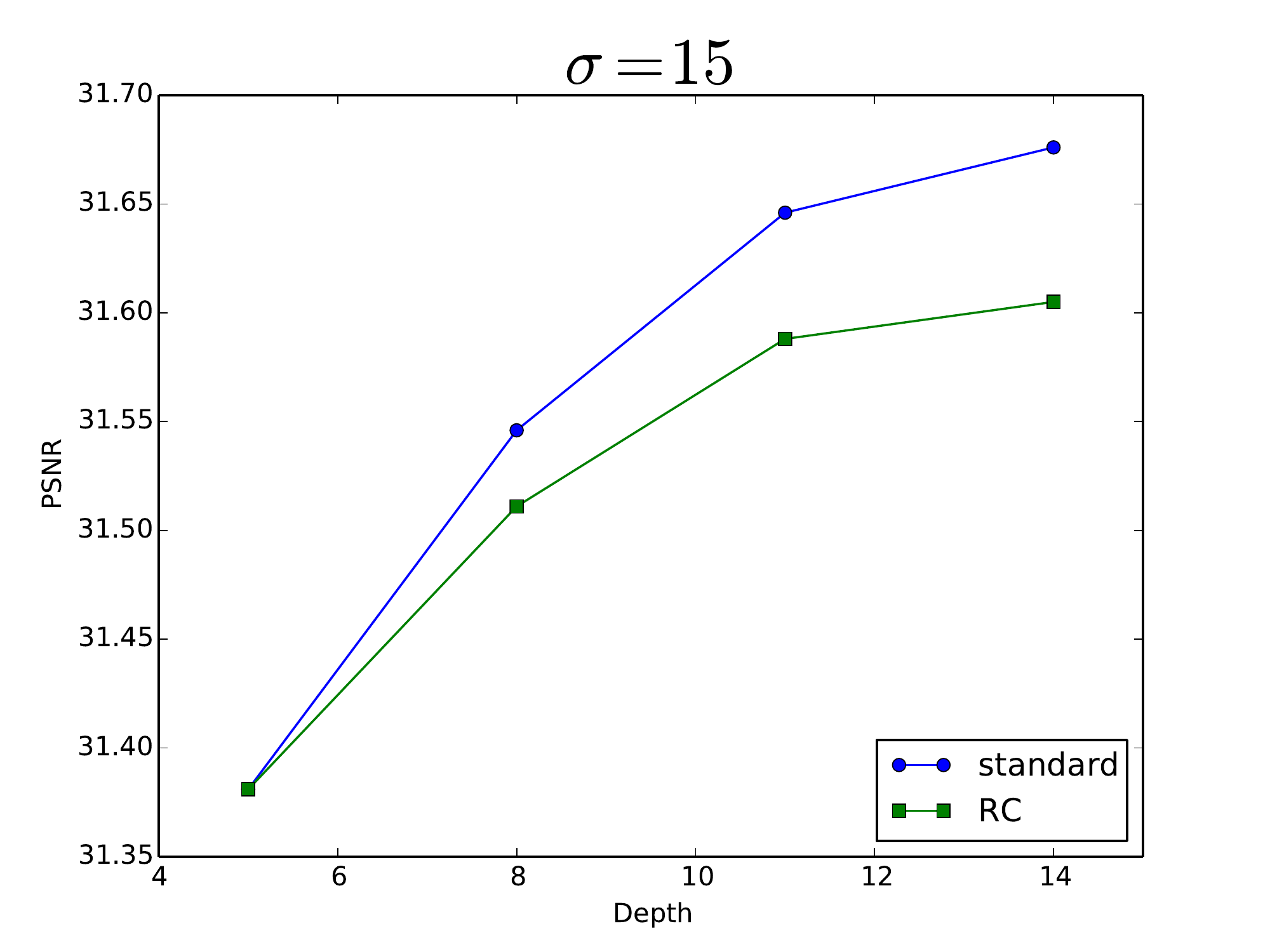}}
  	\subfigure{
  		\includegraphics[width=0.3\textwidth]{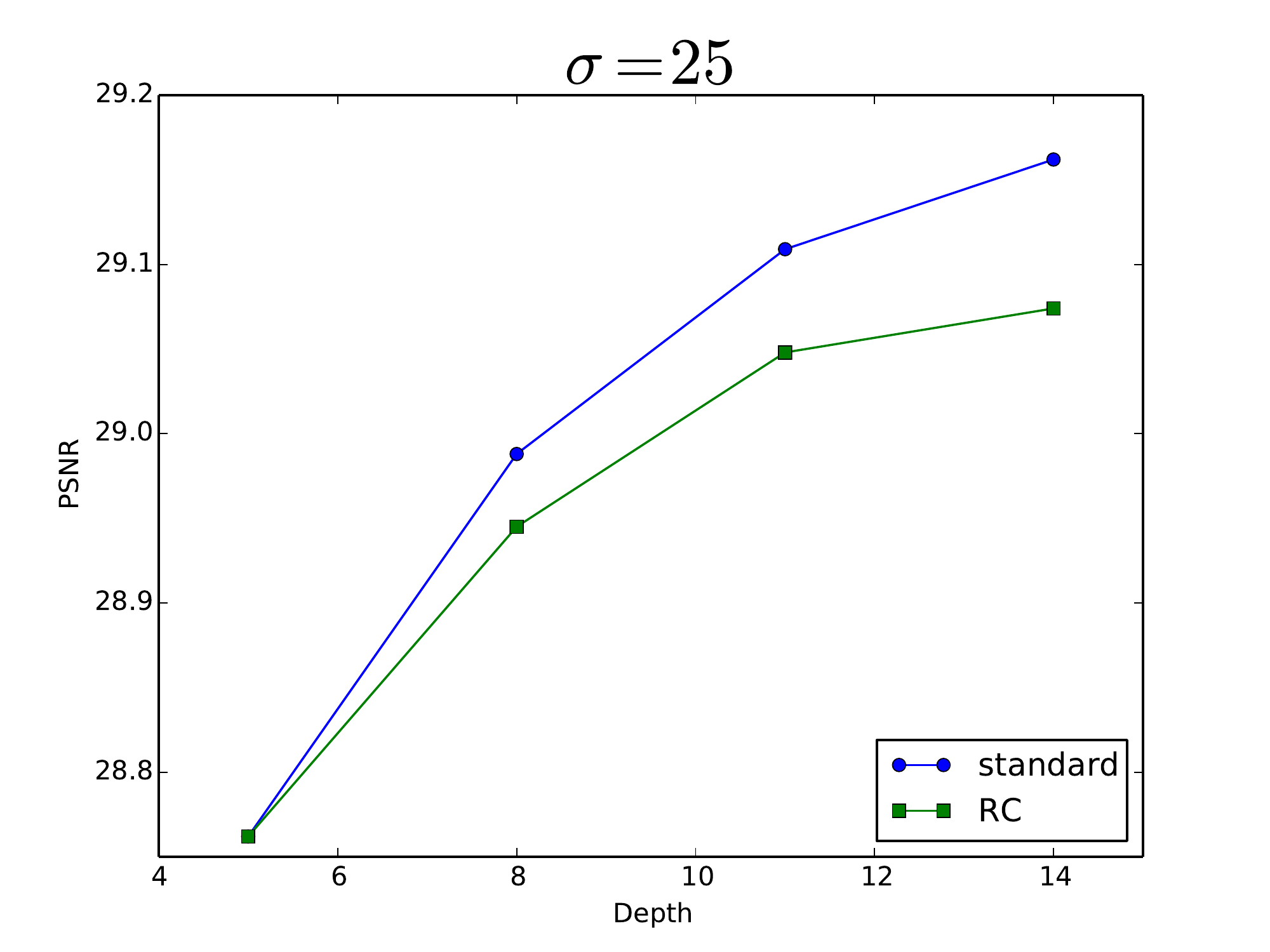}}
  	\subfigure{
  	  		\includegraphics[width=0.3\textwidth]{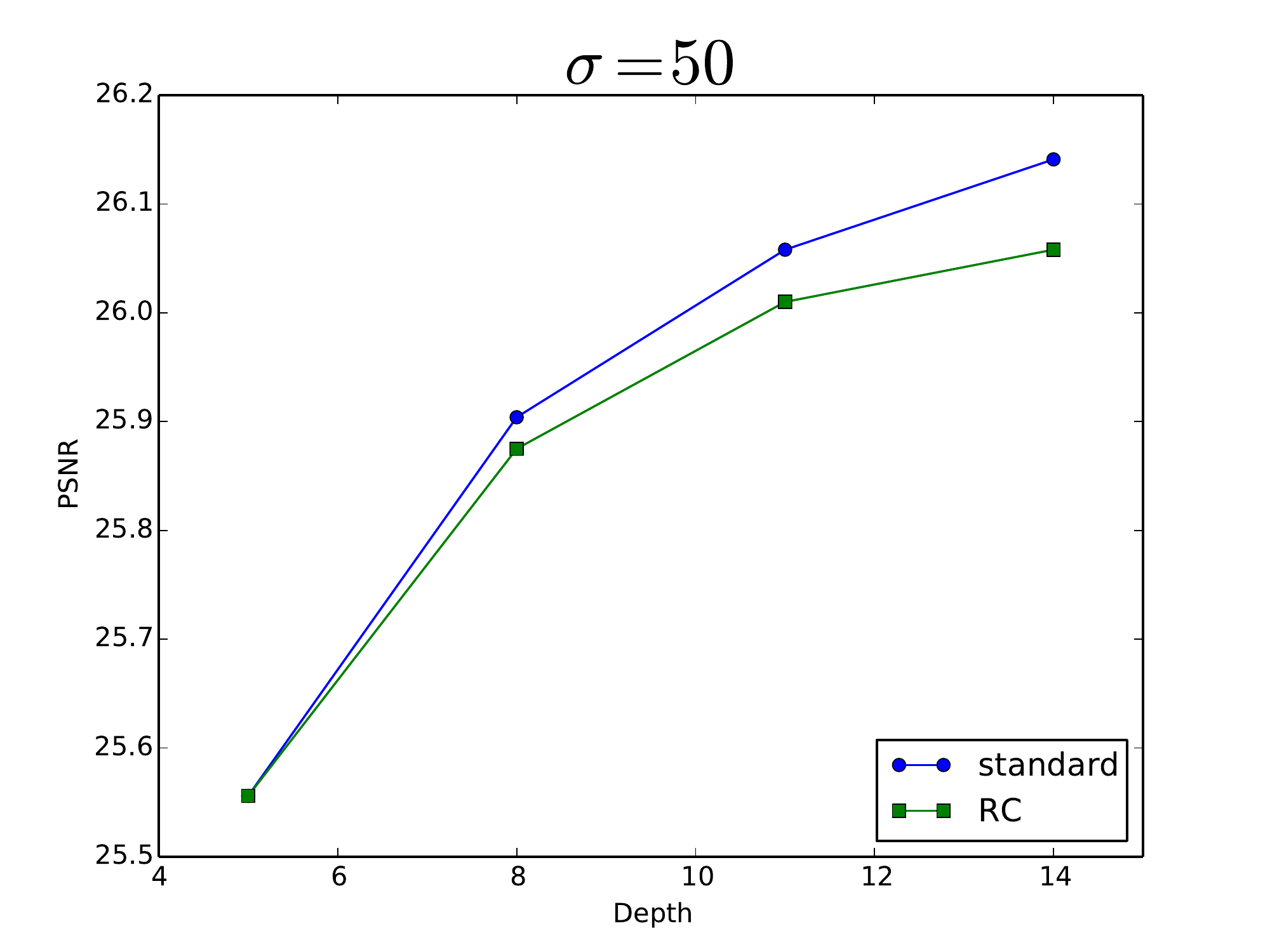}}
  	\caption{We compare the denoise performance in term of PSNR between RC networks and standard networks on BSD. The unrolling step of RC networks ranges from 1 to 4. Results of denoise level 15, 25 and 50 are shown in (a), (b) and (c) respectively.}
  	\label{fig_denoise}
   \end{figure*}
   
   \begin{table}[t]
   	\caption{Comparisons between RC networks and standard networks on CIFAR. `Depth' is the unrolled depth.}
   	\centering
   	\begin{tabular}{c|cc|cc}
   		\hline
   		\hline
   		                  & CIFAR-10 & CIFAR-100 & Parameters & Depth\\
   		\hline
   		$R_2^1$           &   14.46    & 40.30      &  1.259M & 6\\
   		$R_2^2$          &   8.53    & 32.22         & 1.260M & 10\\
   		$R_2^3$          &   8.44    & 31.78        & 1.262M & 14\\
   		$R_2^4$           &   7.65    & 30.33        & 1.263M & 18\\
   		\hline
   		$S_2^2$           &   7.93    & 30.89        & 2.514M & 10\\
   		$S_2^3$           &   7.41    & 30.26        & 3.768M & 14\\
   		$S_2^4$           &   6.86    & 28.35        & 5.023M & 18\\
   		\hline
   		$S_2^4$ $\times 0.5$ & 8.32 & 31.54 & 1.258M & 18\\
   		\hline
   	\end{tabular}
   \label{tab_cifar_18}
   \end{table}
   
\subsection{Is RC Helpful?}
We evaluate whether unrolling a cell multiple steps helpful. We train each network three times, then average the results. Test errors on CIFAR are shown in Tab.~\ref{tab_cifar_18}. PSNR on BSD is shown in Fig.~\ref{fig_denoise}. The performance is consistently improved w.r.t unrolling steps. However, the performance of RC networks can't match their corresponding standard ones. This is not surprised. It is worth noting that $R_2^4$ whose original depth is 6 has significant better performance than $S_2^2$ whose depth is 10. 

Those results indicate that RC is helpful for both semantic level tasks and pixel level tasks. Moreover, the numbers of parameters of $R_2^4$ and $S_2^4$ $\times 0.5$ in Tab.~\ref{tab_cifar_18} are nearly the same. But the former has lower errors. This indicates that RC is valuable for learning compact networks.

\begin{table}[t]
	\caption{Test errors on CIFAR of $R_2^4$ with different BN usages.}
	\centering
	\begin{tabular}{c|ccc}
		\hline
		\hline
		                   & No & Shared & Independent \\
		\hline
		CIFAR-10     & 10.37 & 21.00 & 7.65 \\
		CIFAR-100   & 38.56 & 54.52 & 30.33 \\
		\hline
	\end{tabular}
\label{tab_bn_cifar}
\end{table}

\begin{table}[t]
	\caption{PSNR on BSD of $R_2^4$ with different BN usages. `-' indicates the result is not convergent on test set.}
	\centering
	\begin{tabular}{c|ccc}
		\hline
		\hline
		                   & No & Shared & Independent\\
		\hline
		$\sigma=15$     & 31.36 & - & 31.61 \\
		$\sigma=25$   & 28.78 & - & 29.07 \\
		$\sigma=50$     & 25.77 & 17.54 & 26.06 \\
		\hline

	\end{tabular}
	
\label{tab_bn_denoise}
\end{table}

\subsection{Is Independent BN Helpful?}
We evaluate whether learning independent BN layers is helpful. We compare three different BN usages: without BN, shared BN and independent BN. Results for image classification and image denoise are shown in Tab.~\ref{tab_bn_cifar} and Tab.~\ref{tab_bn_denoise} respectively. Using independent BN layers improves the performance of RC networks by a large margin. Using shared BN layers has much worse performance than without using BN layers. For denoise, the results on test set are even not convergent. Moreover, we show the parameters of the first BN layer of $C_1$ in Fig.~\ref{fig_BN_0}. Those learned parameters vary over unrolling steps.

Those results indicate that independent BN plays a very important role for training RC networks.

\begin{table}[t]
	\centering
	\caption{Test errors on CIFAR of cost-adjustable $R_2$. The boldface numbers indicate the performance gap between fixed step $R_2$.}
	\begin{tabular}{c|ccc}
		\hline
		\hline
		Steps            & 2 & 3 & 4 \\
		\hline
		CIFAR-10     & 8.46(\textbf{-0.07}) & 8.08(\textbf{-0.36}) & 7.87(\textbf{+0.02}) \\
		CIFAR-100   & 31.55(\textbf{-0.34}) & 31.21(\textbf{-0.57}) & 30.67(\textbf{+0.12}) \\
		\hline
	\end{tabular}
\label{tab_any18}
\end{table}

\begin{figure*}[t]
\centering
 \includegraphics[width=0.9\textwidth]{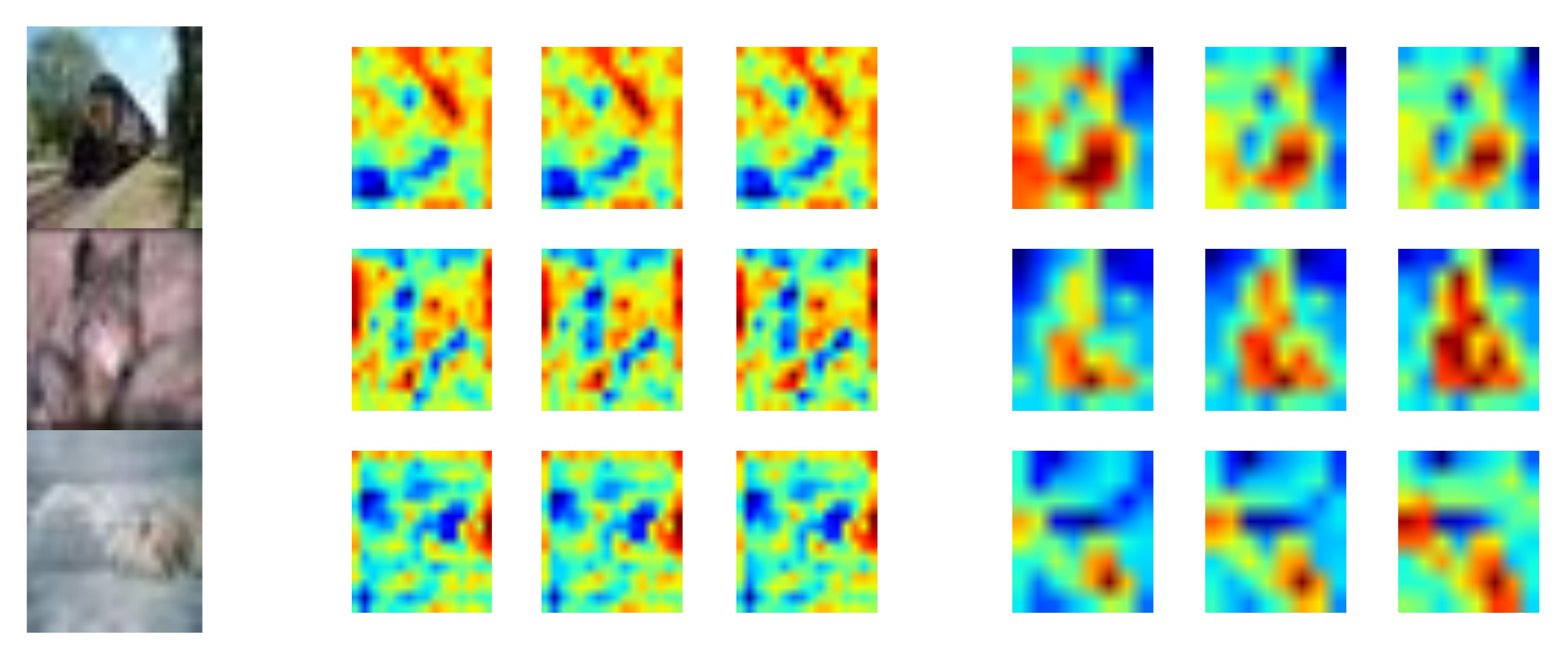}
 \caption{Examples of how features of $R_4$ evolve. The second group of columns are features of $C_2$. The third group of columns are features of $C_3$. Columns of each group correspond to the steps of each cell. Images in the first column are randomly selected from CIFAR-100.}
 \label{fig_features}
\end{figure*}

\begin{figure}[t]
\centering
 \includegraphics[width=0.4\textwidth]{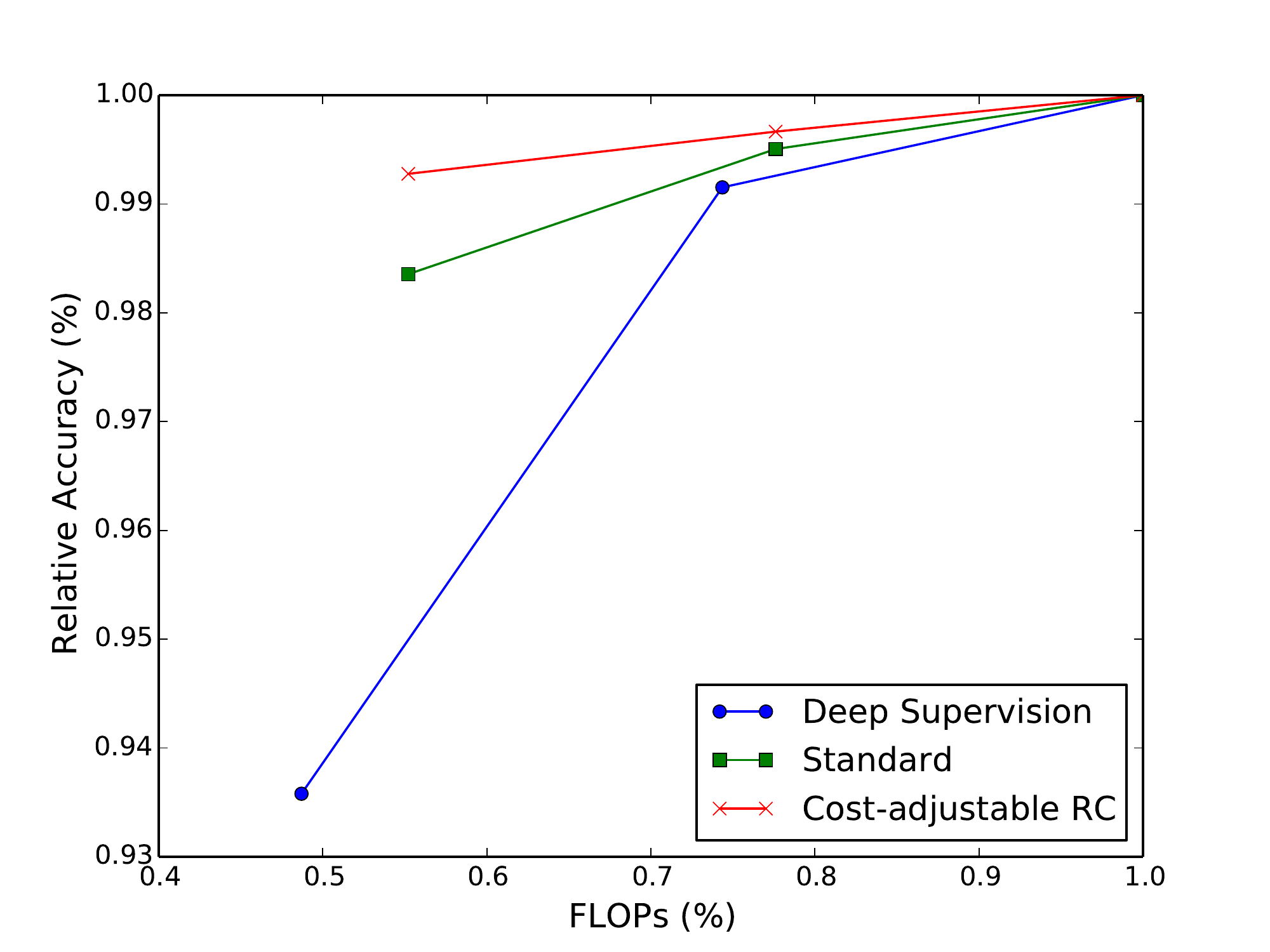}
 \caption{Comparisons of cost-adjustable inference.}
 \label{fig_cost}
\end{figure}
\subsection{Cost-adjustable Training}
\label{sec_53}

\begin{table}[t]
   	\caption{Comparisons between cost-adjustable $R_4$ and standard networks on CIFAR-100. `Depth' is the unrolled depth.}
   	\centering
   	\begin{tabular}{c|ccc}
   		\hline
   		\hline
   		                   & CIFAR-100 & Parameters & Depth\\
   		\hline
   		$R_4^1$          & 25.83      &  11.250M & 18\\
   		$R_4^2$          &  25.54       & 11.250M & 26\\
   		$R_4^3$           & 25.29        & 11.250M & 34\\
   		\hline
   		$S_4^1$           & 26.45        & 11.217M & 18\\
   		$S_4^2$           & 25.59        & 17.488M & 26\\
   		$S_4^3$           & 25.22        & 23.759M & 34\\
   		\hline
   	\end{tabular}
   \label{tab_cifar_34}
   \end{table}
   
We compare cost-adjustable training between fixed step training on CIFAR. The step of the cost-adjustable RC network ranges from 2 to 4 which is sampled from discrete distribution $\{ 0.2, 0.3, 0.5\}$. For the fixed step RC networks, we use the results in Tab.~\ref{tab_cifar_18}. We train cost-adjustable RC networks via Alg.~\ref{alg:B}. As shown in Tab.~\ref{tab_any18}, the overall performance of cost-adjustable RC networks are competitive with the fixed step RC networks which are trained independently. For step 2 and step 3, the former is even better. 

We train a larger network, \ie ResNet-34~\cite{he2016identity} for further comparisons. ResNet-34 has 4 groups of residual blocks. We set the number of blocks of each group to 4 for convenience which is slightly different from original ResNet-34. The first block of each group may change the number of channels. We simply keep those blocks. The rest of blocks in each group have same architectures. We replace those blocks with a single RC block. When each RC block is unrolled with its maximum step, the computational graph of the whole RC network is exactly the same with the standard one. This RC network has the following architecture:
\begin{align}
& T_1 \rightarrow C_1^n(64) \rightarrow T_2 \rightarrow C_2^n(128) \rightarrow \\ \nonumber
& T_3 \rightarrow C_3^n(256) \rightarrow T_4 \rightarrow C_4^n(512)
\end{align}
where $T$ is a non-recurrent block. We denote this network as $R_4$. We show the learned parameters of the fist BN layer of $T_2$ when $C_1$ unrolls different steps from 1 to 3 trained on CIFAR-100 in Fig.~\ref{fig_BN_1}. Those parameters over steps look similar but they are not exactly the same.

We compare our method with the standard networks with varied depth in Tab.~\ref{tab_cifar_34}. We compare our method with~\cite{lee2015deeply-supervised} which trains multiple intermediate classifiers. We insert intermediate classifiers after $C_2^3$ and $C_3^3$ respectively. We show how their relative accuracy varies w.r.t their relative FLOPs. RC network is better than others.

\subsection{Visual Analysis}
We provide visualizing results which help us understanding RC networks. We show denoise results on BSD in Fig.~\ref{fig_imgs}. We can see how noise is recursively removed as the unrolling step grows, \eg the sky region. We show how feature maps evolve over steps for image classification on CIFAR-100. Specifically, given an image, we compute its feature maps at the end of $C_2$ and $C_3$ unrolled by their maximum steps. Then we select the channels with maximum average activations. We show those features in Fig.~\ref{fig_features}. It seems that they vary mildly over steps.

\section{Conclusions}
\label{sec_6}
We have shown that the performance of a network is improved if we unroll its cell multiple steps with independent BN layers. Thus recurrent convolution indirectly plays a role in compact neural network representation by reduction the redundancy across layers. We have shown RC networks can perform cost-adjustable inference by varying its unrolling steps. We propose double independent BN to train cost-adjustable RC networks. We have provided insights on why our method works. We believe RC for compact and cost-adjustable neural networks is a potential direction which is worth to be further explored.

{\small
\bibliographystyle{ieee}
\bibliography{egbib}
}

\end{document}